\documentclass[11pt,a4paper]{article}
\usepackage[hyperref]{acl2019}
\usepackage{times}
\usepackage{latexsym}
\usepackage{amsfonts}
\usepackage{amsmath}
\usepackage{amssymb}
\usepackage{multirow}
\usepackage{booktabs}
\usepackage{graphicx}
\usepackage{url}
\usepackage{hyperref}
\usepackage{color,colortbl}
\definecolor{Gray}{gray}{0.9}

\usepackage[textsize=scriptsize]{todonotes}
\definecolor{blue}{RGB}{0, 93, 170}

\aclfinalcopy % Uncomment this line for the final submission
\def\aclpaperid{2046} %  Enter the acl Paper ID here

\title{Improving Question Answering over Incomplete KBs\\ with Knowledge-Aware Reader}
\author{
 Wenhan Xiong$^\dagger$,
 Mo Yu$^\ast$, 
 Shiyu Chang$^\ast$, 
 Xiaoxiao Guo$^\ast$, 
 William Yang Wang$^\dagger$
\\ 
 $^\dagger$ University of California, Santa Barbara\\
 $^\ast$ IBM Research\\
 \{xwhan, william\}@cs.ucsb.edu, yum@us.ibm.com, \{shiyu.chang, xiaoxiao.guo\}@ibm.com  
 }
\date{}

\begin{document}
\maketitle
\begin{abstract}
We propose a new end-to-end question answering model, which learns to aggregate answer evidence from an incomplete knowledge base (KB) and a set of retrieved text snippets. Under the assumptions that the structured KB is easier to query and the acquired knowledge can help the understanding of unstructured text, our model first accumulates knowledge of entities from a question-related KB subgraph; then reformulates the question in the latent space and reads the texts with the accumulated entity knowledge at hand.  
The evidence from KB and texts are finally aggregated to predict answers. 
On the widely-used KBQA benchmark WebQSP, our model achieves consistent improvements across settings with different extents of KB incompleteness.\footnote{\scriptsize{\url{https://github.com/xwhan/Knowledge-Aware-Reader}.}}

\end{abstract}

%%%%%%%%%%%%%%%%%%%%%%%%%%%%%%%%%%%%%%%%%%%%%%%%%%%%%%%%%%%%%%%%%%%%%%%%%%%%%%%%%%%%%%%%%%
%%%%%%%%%%%%%%%%%%%%%%%%%%%%%%%%%%%%%%%%%%%%%%%%%%%%%%%%%%%%%%%%%%%%%%%%%%%%%%%%%%%%%%%%%%
\section{Introduction}
Knowledge bases (KBs) are considered as an essential resource for answering factoid questions. However, accurately constructing KB with a well-designed and complicated schema requires lots of human efforts, which inevitably limits the coverage of KBs~\cite{min2013distant}. As a matter of fact, KBs are often incomplete and insufficient to cover full evidence required by open-domain questions.

\begin{figure}[t]
\centering
\includegraphics[width=\linewidth]{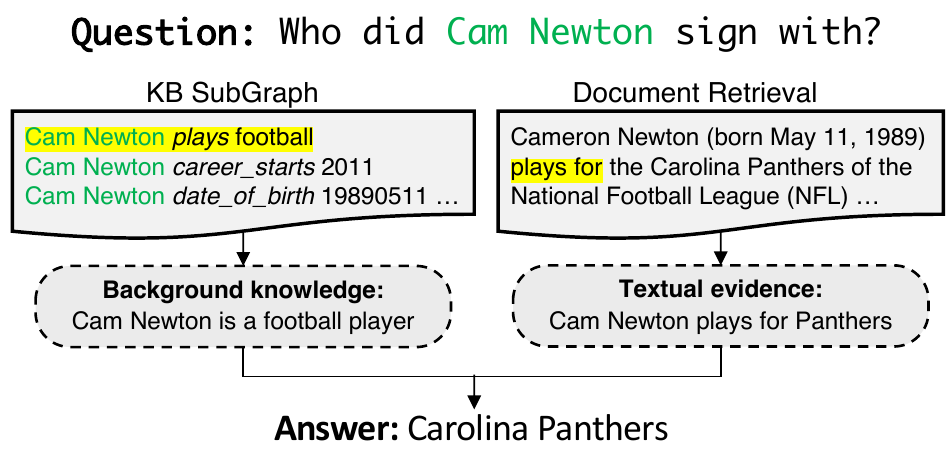}
\caption{\small{A real example from WebQSP. Here the answer cannot be directly found in the KB. But the knowledge provided by the KB, \emph{i.e.}, \textit{{Cam Newton is a football player}}, indicates he \textbf{signed with} \textit{the team he plays for}.} This knowledge can be essential for recognizing the relevant text piece.}
\vspace{-0.2 in}
\label{system}
\end{figure}

On the other hand, the vast amount of unstructured text on the Internet can easily cover a wide range of evolving knowledge, which is commonly used for open-domain question answering~\cite{chen2017reading, wang2018r}. Therefore, to improve the coverage of KBs, it is straightforward to augment KB with text data. %  On the other hand, 
Recently, text-based QA models along~\cite{seo2016bidirectional,xiong2017dcn+,yu2018qanet} have achieved remarkable performance when dealing with a single passage that is guaranteed to include the answer.  However, they are still insufficient when multiple documents are presented. We hypothesize this is partially due to the lack of background knowledge while distinguishing relevant information from irrelevant ones (see Figure~\ref{system} for a real example).  

% To make better use of 
To better utilize textual evidence for improving QA over incomplete KBs, this paper presents a new end-to-end model, which consists of (1) a simple yet effective subgraph reader that accumulates knowledge of each KB entity from a question-related KB subgraph; and (2) a knowledge-aware text reader that selectively incorporates the learned KB knowledge about entities with a novel conditional gating mechanism. 
With the specifically designed gate functions, our model has the ability to dynamically determine how much KB knowledge to incorporate while encoding questions and passages, thus is able to make the structured knowledge more compatible with the text information.
Compared to the previous state-of-the-art~\cite{sun2018open}, our model achieves consistent improvements with a much more efficient pipeline, which only requires a single pass of the evidence resources. 

\begin{figure*}[th]
\centering
\includegraphics[width=\linewidth]{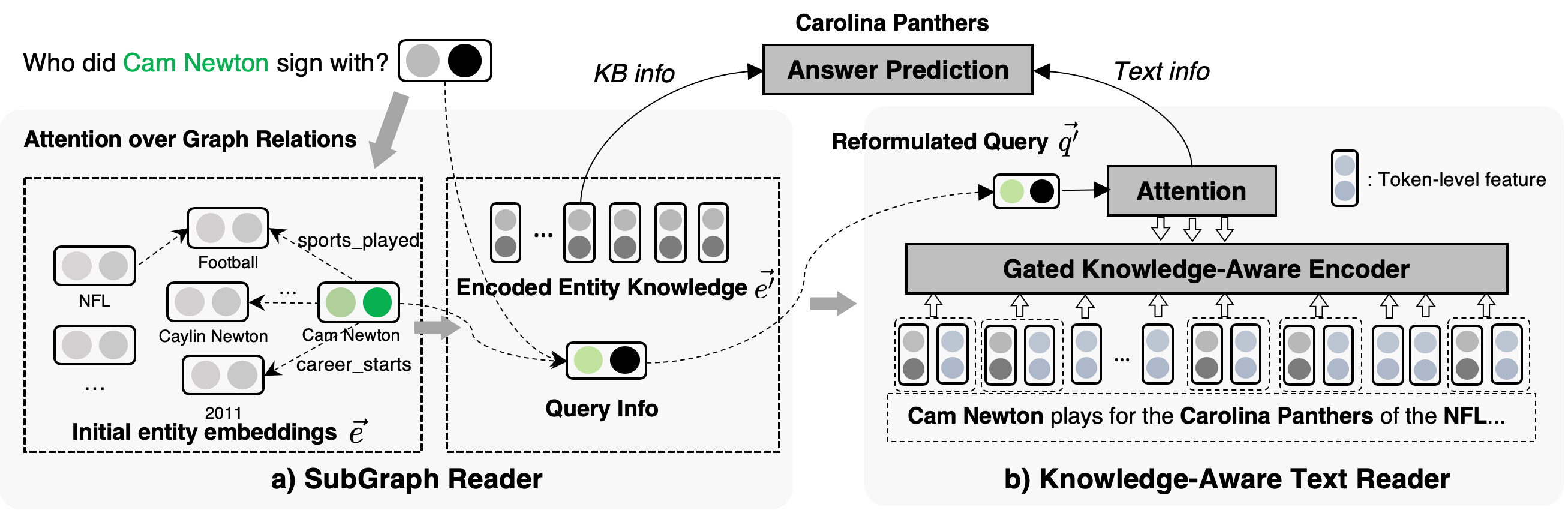}
\caption{\small{Model Overview. The subgraph reader \textbf{a}) first utilizes graph attention networks~\cite{velivckovic2017graph} to collect information for each entity in the question-related subgraph. The learned knowledge of each entity ($\vec{e'}$) is then passed to the text reader \textbf{b)} to reformulate the question representation ($\vec{q'}$) and encode the passage in a knowledge-aware manner.  Finally, the information from the text and the KB subgraph is aggregated for answer entity prediction.}}
\label{model}
\vspace{-0.15 in}
\end{figure*}
\section{Task Definition}

The QA task we consider here requires answering questions by reading knowledge base tuples $\mathcal{K} = \{(e_s, r, e_o)\}$ and retrieved Wikipedia documents $\mathcal{D}$. To build a scalable system, we follow \citet{sun2018open} and only consider a subgraph for each question. The subgraph is retrieved by running Personalized PageRank~\cite{haveliwala2002topic} from the topic entities\footnote{Annotated by STAGG~\cite{yih2014semantic}.} (entities mentioned by the question: $\mathcal{E}_0 = \{e|e\in Q\}$).  The documents $\mathcal{D}$ are retrieved by an existing document retriever~\cite{chen2017reading} and further ranked by Lucene index.
% \footnote{\url{http://lucene.apache.org/}}. 
The entities in documents are also annotated and linked to KB entities.  For each question, the model tries to retrieve answer entities from a candidate set including all KB and document entities.

%%%%%%%%%%%%%%%%%%%%%%%%%%%%%%%%%%%%%%%%%%%%%%%%%%%%%%%%%%%%%%%%%%%%%%%%%%%%%%%%%%%%%%%%%%
\section{Model}
The core components of our model consist of a graph-attention based KB reader~(\S\ref{ssec:kb_reader}) and a knowledge-aware text reader~(\S\ref{ssec:text_reader}). The interaction between the modules is shown in Figure~\ref{model}.

%%%%%%%%%%%%%%%%%%%%%%%%%%%%%%%%%%%%%%%%%%%%%%%%%%%%%%%%%%%%%%%%%%%%%%%%%%%%%%%%%%%%%%%%%%
\subsection{\underline{S}ub\underline{G}raph Reader}
\label{ssec:kb_reader}
This section describes the KB subgraph reader (\textsc{SGReader}), which employs graph-attention techniques to accumulate knowledge of each subgraph entity ($e$) from its linked neighbors ($N_e$).  The graph attention mechanism is particularly designed to take into account two important aspects: (1) whether the neighbor relation is relevant to the question; (2) whether the neighbor entity is a topic entity mentioned by the question. After the propagation, the \textsc{SGReader} finally outputs a vectorized representation for each entity, encoding the knowledge indicated by its linked neighbors.
% how its neighbors in KB match the question \wh{need discussion} \shiyu{discuss}.

%%%%%%%%%%%%%%%%%%%%%%%%%%%%%%%%%%%%%%
\noindent \textbf{Question-Relation Matching}$\quad$
To match the question and KB relation in an isomorphic latent space, we apply a shared LSTM to encode the question $\{w^{q}_1, w^{q}_2, ..., w^{q}_{l_q}\}$ and the tokenized relation $\{w^{r}_1, w^{r}_2, ..., w^{r}_{l_r}\}$. With the derived hidden states $\mathbf{h^q} \in \mathbb{R}^{l_q \times d_h}$ and $\mathbf{h^r} \in \mathbb{R}^{l_r \times d_h}$ for each word, we first compute the representation of  relations with a self-attentive encoder:
\begin{equation}
\small
\begin{aligned}
    \vec{r} = \sum_{i}\alpha_i \vec{h}^r_{i},~ \alpha_i \propto \exp(\vec{w_r} \cdot \vec{h}^r_{i} ), \nonumber
\end{aligned}
\end{equation}
where $\vec{h}^r_{i}$ is the $i$-th row of $\mathbf{h^r}$ and $\vec{w_r}$ is a trainable vector.  Since a question needs to be matched with different relations and each relation is only described by part of the question, instead of matching the relations with a single question vector, we calculate the matching score in a more fine-grained way. Specifically, we first use $\vec{r}$ to attend each question token and then model the matching  $s_r$ by a dot product as follows:
\begin{equation}
\small
\begin{aligned}
    s_{r} = \vec{r} \cdot \sum_{j}\beta_j\vec{h}^q_{j}, \beta_j \propto \exp(\vec{r} \cdot \vec{h}^q_{j}). \nonumber
\end{aligned}
\end{equation}

%%%%%%%%%%%%%%%%%%%%%%%%%%%%%%%%%%%%%%
\paragraph{Extra Attention over Topic Entity Neighbors}

In addition to the question-relation similarities, we find another binary indicator feature derived from the topic entity is very useful.  This indicator is defined as $I[e_i \in \mathcal{E}_0]$ for a neighbor $(r_i, e_i)$ of an arbitrary entity $e$. Intuitively, if one neighbor links to a topic entity that appear in the question then the corresponding tuple $(e, r_i, e_i)$ could be more relevant than other non-topic neighbors for question answering. Formally, the final attention score $\tilde{s}_{(r_i, e_i)}$ over each neighbor $(r_i, e_i)$ is defined as:
\begin{equation}
\small
\begin{aligned}
  \tilde{s}_{(r_i, e_i)} \propto \exp( I[e_i \in \mathcal{E}_0] + s_{r_i} ). \nonumber
\end{aligned}
\end{equation}
% where $c$ is a hyperparameter. We use $c = 1$ throughout our experiments.

%%%%%%%%%%%%%%%%%%%%%%%%%%%%%%%%%%%%%%
\noindent \textbf{Information Propagation from Neighbors}
To accumulate the knowledge from the linked tuples, we define the propagation rule for each entity $e$:
\begin{equation}
\small
\begin{aligned}
\vec{e'} = \gamma^e \vec{e} + (1 - \gamma^e) \sum_{(e_i, r_i)\in N_e} \tilde{s}_{(r_i, e_i)} \sigma(\mathbf{W_e}[\vec{r_i};\vec{e_i}]), \nonumber
\end{aligned}
\end{equation}
where $\vec{e}$ and $\vec{e_i}$ are pre-computed knowledge graph embeddings, $\mathbf{W_e} \in \mathbb{R}^{h_d \times 2h_d}$ is a trainable transformation matrix and $\sigma(\cdot)$ is an activation function.  In addition, $\gamma_e$ is a trade-off parameter calculated by a linear gate function as $\gamma^e = g(\vec{e}, \sum_{(e_i, r_i)\in N_e} \tilde{s}_{(r_i,e_i)} \sigma(\mathbf{W_e}[\vec{r_i};\vec{e_i}]))$\footnote{$g(x,y) = \textrm{sigmoid}(\mathbf{W}[x;y]) \in (0,1)$.} , which controls how much information in the original entity representation should be retained.\footnote{The above step can be viewed as a gated version of the graph encoding techniques in NLP, e.g., \cite{song2018graph,xu2018graph2seq}.
These general graph-encoders and graph-attention techniques may help when the questions require more hops and we leave the investigation to future work.}

%%%%%%%%%%%%%%%%%%%%%%%%%%%%%%%%%%%%%%%%%%%%%%%%%%%%%%%%%%%%%%%%%%%%%%%%%%%%%%%%%%%%%%%%%%
%%%%%%%%%%%%%%%%%%%%%%%%%%%%%%%%%%%%%%%%%%%%%%%%%%%%%%%%%%%%%%%%%%%%%%%%%%%%%%%%%%%%%%%%%%
%%%%%%%%%%%%%%%%%%%%%%%%%%%%%%%%%%%%%%%%%%%%%%%%%%%%%%%%%%%%%%%%%%%%%%%%%%%%%%%%%%%%%%%%%%
\subsection{\underline{K}nowledge-\underline{A}ware Text Reader}
\label{ssec:text_reader}
With the learned KB embeddings, %we now describe how to 
our model enhances text reading with~\textsc{KAReader}.  Briefly, we use an existing reading comprehension model~\cite{chen2017reading} and improve it by learning more knowledge-aware representations for both question and documents. 
%%%%%%%%%%%%%%%%%%%%%%%%%%%%%%%%%%%%%%
\paragraph{Query Reformulation in Latent Space}
First, we update the question representation in a way that the KB knowledge of the topic entity can be incorporated. This allows the reader to discriminate relevant information beyond text matching. 

Formally, we first take the original question encoding $\mathbf{h^q}$ and apply a self-attentive encoder to get a stand-alone question representation: $\vec{q} = \sum_{i}b_i\vec{h}^q_{i}$. We collect the topic entity knowledge of the question by $\vec{e^q} = \sum_{e \in \mathcal{E}_0} \vec{e'} / \vert\mathcal{E}_0 \vert$. Then we apply a gating mechanism to fuse the original question representation and the KB knowledge:
\begin{equation}
\small
\begin{aligned}
    \vec{q'} = \gamma^q \vec{q} + (1-\gamma^q)\tanh(\mathbf{W^q}[\vec{q}, \vec{e^q}, \vec{q} - \vec{e^q}]), \nonumber
\end{aligned}
\end{equation}
where $\mathbf{W^q} \in \mathbb{R}^{h_d \times 3h_d}$, and $\gamma^q = \textrm{sigmoid}(\mathbf{W^{gq}}[\vec{q}, \vec{e^q}, \vec{q} - \vec{e^q}])$ is a linear gate. 

%%%%%%%%%%%%%%%%%%%%%%%%%%%%%%%%%%%%%%
\paragraph{Knowledge-aware Passage Enhancement}
 To encode the retrieved passages, we use a standard bi-LSTM, which takes several token-level features\footnote{We use the same set of features as in~\cite{chen2017reading} except for the tagging labels.}. With the entity linking annotations in passages, we fuse the entity knowledge with the token-level features in a similar fashion as the query reformulation process. However, instead of applying a standard gating mechanism~\cite{yang2017leveraging,mihaylov2018knowledgeable}, we propose a new conditional gating function that explicitly conditions on the question $\vec{q'}$.  This simple modification allows the reader to dynamically select the inputs according to their relevance to the question.  Considering a passage token $w^{d}_i$ with its token features $\vec{f}^{d}_{w_i}$ and its linked entity $e_{w_i}$\footnote{Non-entity tokens are encoded with token-level features only.}, we define the conditional gating function as:
\begin{align*}
% \tiny
\small
\begin{aligned}
    \vec{i}^{d}_{w_i} &= \gamma^d \vec{e'}_{w_i} + (1 - \gamma^d)\vec{f}^{d}_{w_i},
    \text{  where}\\
    \gamma^d &= \textrm{sigmoid}(\mathbf{W^{gd}}[\vec{q}\cdot \vec{e'}_{w_i};\vec{q}\cdot \vec{f}^{d}_{w_i}]).
\end{aligned}
\end{align*}
% 
% \end{equation*}
$\vec{e'}_{w_i}$ denotes the entity embedding learned by our \textsc{SGReader}. 

%%%%%%%%%%%%%%%%%%%%%%%%%%%%%%%%%%%%%%
\paragraph{Entity Info Aggregation from Text Reading}
Finally we feed the knowledge-augmented inputs $\vec{i}^{d}_{w_i}$ into the biLSTM and use the output token-level hidden state $\vec{h}^{d}_{w_i}$ to calculate the attention scores $\lambda_i=\vec{q'}^T \vec{h}^{d}_{w_i}$. Afterwards, we get each document's representation as $\vec{d} = \sum_i \lambda_i \vec{h}^{d}_{w_i}$.  For a certain entity $e$ and all the documents containing $e$: $\mathcal{D}^e = \{d|e \in d\}$, we simply aggregate the information by averaging the representations of linked documents as $\vec{e}_d = \frac{1}{|\mathcal{D}^e|}\sum_{d \in \mathcal{D}^e} \vec{d}$.

%%%%%%%%%%%%%%%%%%%%%%%%%%%%%%%%%%%%%%%%%%%%%%%%%%%%%%%%%%%%%%%%%%%%%%%%%%%%%%%%%%%%%%%%%%
\subsection{Answer Prediction}
With entities representations ($\vec{e'}$ and $\vec{e^d}$), we predict the probability of an entity being the answer by matching the query vectors and the entity representations: $s^e = \sigma_s(\vec{q'}^T \mathbf{W_s}[\vec{e'};\vec{e^d}])$.

\section{Experiment}
%%%%%%%%%%%%%%%%%%%%%%%%%%%%%%%%%%%%%%%%%%%%%%%%%%%%%%%%%%%%%%%%%%%%%%%%%%%%%%%%%%%%%%%%%%

\begin{table*}[!ht]
    \centering
    \small
    \begin{tabular}{l|cc|cc|cc|cc}
    \toprule
        \multirow{2}{0.8cm}{\textbf{Model}} & \multicolumn{2}{c|}{\textbf{10\% KB}} & \multicolumn{2}{c|}{\textbf{30\% KB}} & \multicolumn{2}{c|}{\textbf{50\% KB}} &\multicolumn{2}{c}{\textbf{100\% KB}}\\
        & Hit@1 & F1 & Hit@1 & F1 & Hit@1 & F1 & Hit@1 & F1 \\
        \cmidrule{1-9} 
        \textsc{KV-KB} & 12.5 & 4.3 & 25.8 & 13.8 & 33.3 & 21.3 & 46.7 & 38.6 \\
        \textsc{GN-KB} &15.5 & 6.5 & 34.9 & \textbf{20.4} & 47.7 & \textbf{34.3} & \textbf{66.7} & \textbf{62.4}  \\
        \textsc{SGReader} (Ours) & \textbf{17.1} & \textbf{7.0} & \textbf{35.9} & 20.2 & \textbf{49.2} & 33.5 & 66.5 & 58.0\\
        \cmidrule{1-9}
        \textsc{KV-KB+Text} & 24.6 & 14.4 & 27.0 & 17.7 & 32.5 & 23.6 & 40.5 & 30.9  \\
        \textsc{GN-LF} & 29.8 & 17.0 & 39.1 & 25.9 & 46.2 & 35.6 & 65.4 & 56.8 \\
        \textsc{GN-EF} & 31.5 & 17.7 & 40.7 & 25.2 & 49.9 & 34.7 & 67.8 & 60.4 \\
        % \rowcolor{Gray}
        % \textsc{GN-EF+LF} & 33.3 & 19.3 & 42.5 & 26.7 & 52.3 & 37.4 & 68.7 & 62.3  \\
        \textsc{SGReader} + \textsc{KAReader} (Ours) & \textbf{33.6} & \textbf{18.9} & \textbf{42.6} & \textbf{27.1} &\textbf{52.7} & \textbf{36.1} & 67.2 & 57.3\\
        \midrule
        \rowcolor{Gray}
        \textsc{GN-LF+EF}~(ensemble)& 33.3 & 19.3 & 42.5 & 26.7 & 52.3 & 37.4 & 68.7 & 62.3\\
        \bottomrule
    \end{tabular}
    \vspace{-0.05 in}
    \caption{\small Comparisons with Key-Value Memory Networks and GRAFT-Nets under different KB settings.}
    \label{tab:main_results}
    \vspace{-0.1in}
\end{table*}

%%%%%%%%%%%%%%%%%%%%%%%%%%%%%%%%%%%%%%%%%%%%%%%%%%%%%%%%%%%%%%%%%%%%%%%%%%%%%%%%%%%%%%%%%%
\subsection{Setup}
\paragraph{Dataset} 
Our experiments are based on the WebQSP dataset~\cite{yih2016value}. To simulate the real-world scenarios, we test our models following the settings of~\cite{sun2018open}, where the KB is downsampled to different extents.  For a fair comparison, the retrieved document set is the same as the previous work.

% \footnote{Here we do not consider the WikiMovies~\cite{miller2016key} benchmark used by \citet{sun2018open}, as it does not use real-world KBs and includes only synthetic questions with a limited vocabulary. These single-domain questions make document retrieval easier and do not reflect the real challenge of text reading over retrieved text.}

%%%%%%%%%%%%%%%%%%%%%%%%%%%%%%%%%%%
% \vspace{0.05in}
\noindent \textbf{Baselines and Evaluation} $\quad$
Key-Value (KV) Memory Network~\cite{miller2016key} is a simple baseline that treats KB triples and documents as memory cells.  Specifically, we consider its two variants, \textbf{KV-KB}  and \textbf{KV-KB+Text}.  The former is a KB-only model while the latter uses both KB and text.  We also compare to the latest method \textbf{GraftNet} (GN)~\cite{sun2018open}, which treats documents as a special genre of nodes in KBs and utilizes graph convolution~\cite{kipf2016semi} to aggregate the information. Similar to the KV-based baselines, we denote \textbf{GN-KB} as the KB-only version.  Further, both \textbf{GN-LF} (late fusion) and \textbf{GN-EF} (early fusion) consider both KB and text.  The former one considers KB and texts as two separate graphs, and then ensembles the answer scores.   GN-EF is the existing best single model, which considers KB and texts as a single heterogeneous graph and aggregate the evidence to predict a single answer score for each entity. F1 and His@1 are used for evaluation since multiple correct answers are possible.

\paragraph{Implementation Details}
Throughout our experiments, we use the 300-dimension GloVe embeddings trained on the Common Crawl corpus. The hidden dimension of LSTM and the dimension of entity embeddings are both 100. We use the same pre-trained entity embeddings as used by~\citet{sun2018open}. For graph attention over the KB sub-graph, we limit the max number of neighbors for each entity to be 50. We use the norm for gradient clipping as 1.0. We apply dropout=0.2 on both word embeddings and LSTM hidden states. The max question length is set to 10 and the max document length is set to 50. For optimization, we apply label smoothing with a factor of 0.1 on the binary cross-entropy loss. During training, we use the Adam with a learning rate of 0.001.
%%%%%%%%%%%%%%%%%%%%%%%%%%%%%%%%%%%%%%%%%%%%%%%%%%%%%%%%%%%%%%%%%%%%%%%%%%%%%%%%%%%%%%%%%%
\subsection{Results and Analysis}
We show the main results of different incomplete KB settings in Table~\ref{tab:main_results}. For reference, we also show the results under full KB settings (\emph{i.e.}, 100\%, all of the required evidence is covered by KB). The row of \textsc{SGReader} shows the results of our model using only KB evidence.  Compared to the previous KBQA methods (KV-KB and GN-KB), \textsc{SGReader} achieves better results in incomplete KB settings and competitive performance with the full KB. Here we do not compare with existing methods that utilize semantic parsing annotations~\cite{yih2016value,yu2017improved}.  It is worth noting that \textsc{SGReader} only needs one hop of graph propagation while the compared methods typically require multiple hops.

Augmenting the \textsc{SGReader} with our knowledge-aware reader (\textsc{KAReader}) results in consistent improvements in the settings with incomplete KBs.  Compared to other baselines, although our model is built upon a stronger KB-QA base model, it achieves the largest absolute improvement.  It is worth mentioning that our model is still a single model, but it achieves competitive results to the existing ensemble model (GN-LF+EF). The results demonstrate the advantage of our knowledge-aware text reader. 
% We will give qualitative analysis of our \textsc{KAReader} later in this section.

\begin{table}[t]
    \centering
    \small
    \begin{tabular}{l|cc}
    \toprule
        \textbf{Model} & \textbf{Hit@1} & \textbf{F1}\\
        \midrule
        Full Model & 46.8 & 28.1 \\ 
        % - w/o \textsc{SGReader} & &  \\
        % - w/o \textsc{KAR}  & 38.8 & 20.8 \\
        \midrule
        - w/o query reformulation & 44.4 & 27.6 \\
        - w/o knowledge enhancement & 45.2 & 27.0 \\
        - w/o conditional knowledge gate & 44.4 & 27.0\\
        \bottomrule
    \end{tabular}
    \caption{\small Ablation on dev under the 30\% KB setting.}
    \label{tab:ablation}
    \vspace{-0.2 in}
\end{table}

% \begin{table}[t]
%     \centering
%     \small
%     \begin{tabular}{l|c}
%     \toprule
%         \textbf{Method}  & \textbf{F1}\\
%         \midrule
%         % Full Model & 46.8 & 28.1 \\ 
%         % - w/o \textsc{SGReader} & &  \\
%         % - w/o \textsc{KAR}  & 38.8 & 20.8 \\
%         % \midrule
%         \multicolumn{2}{l}{\textbf{SP Based}} \\
%         STAGG (Yih et al., 2015)
%   & 66.8 \\
%         NSM (Liang et al., 2017) & 69.0 \\
%         NSM w/o pseudo-gold parses & \underline{47.8}\\
%         \midrule
%         \multicolumn{2}{l}{\textbf{Neural Matching}} \\
%         Key-Value Memory Net (Miller et al. 2016) & 38.6\\
%         GraftNet (Sun et al. 2018) & 62.8\\
%         \bottomrule
%     \end{tabular}
%     \caption{Ablation on dev under the 30\% KB setting.}
%     \label{tab:ablation}
% \end{table}

%%%%%%%%%%%%%%%%%%%%%%%%%%%%%%%%%%%

\begin{table*}[t]
    \centering
    \small
    \begin{tabular}{c|l}
    \toprule
    
            1)& Question: Which airport to fly into Rome?\\
         & Groundtruth: \textbf{Leonardo da Vinci-Fiumicino Airport (fb:m.01ky5r), Ciampino-G. B. Pastine International} \\ 
         & \textbf{ Airport (fb:m.033\_52)} \\
         & \underline{\textsc{SGReader}}: Italian Met Office Airport (fb:m.04fngkc) \\
         & \underline{\textsc{SGReader} + \textsc{KAReader}}: \textbf{Leonardo da Vinci-Fiumicino Airport (fb:m.01ky5r)}\\ 
         & Missing knowledge of the incomplete KB: No airport info about Rome.\\
         \midrule
       1) & Question: Where did George Herbert Walker Bush go to college?\\
         & Groundtruth: \textbf{Yale (fb:m.08815)}\\
         & \underline{\textsc{SGReader}}: United States of America (fb:m.09c7w0) \\
         & \underline{\textsc{SGReader} + \textsc{KAReader}}:  \textbf{Yale (fb:m.08815)}\\ 
         & Missing knowledge of the incomplete KB: No college info about George Herbert Walker Bush.\\
         \midrule
    
            2)  & Question: When did Juventus win the champions league?\\
         & Groundtruth: \textbf{1996 UEFA Champions League Final (fb:m.02pt\_57)} \\
         & \underline{\textsc{SGReader}}: 1996 UEFA Super Cup (fb:m.02rw0yt)\\
         & \underline{\textsc{SGReader} + \textsc{KAReader}}: \textbf{1996 UEFA Champions League Final (fb:m.02pt\_57)}\\ 
         & Missing knowledge of the incomplete KB: UEFA Super Cup is not UEFA Champions League Final (fb:m.05nblxt)\\
         \midrule
         
        2)  & Question: What college did Albert Einstein go to?\\
         & Groundtruth: \textbf{ETH Zurich (fb:m.01dyk8), University of Zurich (fb:m.01tpvt)} \\
         & \underline{\textsc{SGReader}}:  Sri Krishnaswamy matriculation higher secondary school (fb:m.0127vh33)\\
         & \underline{\textsc{SGReader} + \textsc{KAReader}}: \textbf{ETH Zurich (fb:m.01dyk8)}\\ 
         & Missing knowledge of the incomplete KB: the answer should be a college (fb:m.01y2hnl)\\
        \midrule
        3) & Question: When is the last time the Denver Broncos won the Superbowl?\\
        & Groundtruth: \textbf{Super Bowl XXXIII (fb:m.076y0)} \\
        & \underline{\textsc{SGReader}}:  \textbf{Super Bowl XXXIII (fb:m.076y0)}\\
         & \underline{\textsc{SGReader} + \textsc{KAReader}}: 1999 AFC Championship game (fb:m.0100z7bp)\\ 
         \midrule
        3) & Question: What was Lebron James first team?\\
        & Groundtruth: \textbf{Cleveland Cavaliers (fb:m.0jm7n)} \\
        & \underline{\textsc{SGReader}}:  \textbf{Cleveland Cavaliers (fb:m.0jm7n)}\\
         & \underline{\textsc{SGReader} + \textsc{KAReader}}: Toronto Raptors (fb:m.0jmcb)\\ 
         
        \bottomrule
    \end{tabular}
    \caption{\small Human analysis on test samples in the 30\% KB settingo. 1) and 2) show some typical examples of the case (83.2\% of all test samples) where the  \textsc{KAReader} improves upon our \textsc{SGReader}. 3) shows some examples where using KB alone is better than using both KB and Text (16.8\%). The Freebase IDs of the entities are also included for reference. }
    \label{tab:human_analysis}
    \vspace{-0.2in}
\end{table*}
\paragraph{Ablation Study} To study the effect of each \textsc{KAReader} component, we conduct ablation analysis under the 30\% KB setting (Table~\ref{tab:ablation}). We see that both query reformulation and knowledge enhancement are essential to the performance. Additionally, we find the conditional gating mechanism proposed in \S\ref{ssec:text_reader} is important. When replacing it with a standard gate function (see the row \emph{w/o conditional knowledge gate}), the performance is even lower than the reader without knowledge enhancement, suggesting our proposed new gate function is crucial for the success of knowledge-aware text reading.  The potential reason is that without the question information, the gating mechanism might introduce some irrelevant and misleading knowledge.

\paragraph{Qualitative Analysis}

In Table~\ref{tab:human_analysis}, there are two major categories of questions that can be better answered using our full model. In the first category, indicated by 1), the answer fact is missing in the KB, mainly because there are no links from the question entities to the answer entity.
In these cases, the \textsc{SGReader} sometimes can predict an answer with a correct type, but the answers are mostly irrelevant to the question. 

The second category, denoted as 2), indicates examples where the KB provides relevant information but does not cover some of the constraints on answers' properties (\emph{e.g.}, answers' entity types).
In the two examples shown above, we can see that \textsc{SGReader} is able to give some reasonable answers but the answers do not satisfy the constraints indicated by the question.

Finally, when the KB is sufficient to answer a question, there are some cases where the \textsc{KAReader} introduces wrong answers into the top-ranked answer list.
We list two examples at the bottom of the Table~\ref{tab:human_analysis}.
These newly included incorrect answers are usually relevant to the original questions but come from the noises in machine reading.
These cases suggest that our concatenation-based knowledge aggregation still has some room for improvement, which we leave for future work.

%%%%%%%%%%%%%%%%%%%%%%%%%%%%%%%%%%%%%%%%%%%%%%%%%%%%%%%%%%%%%%%%%%%%%%%%%%%%%%%%%%%%%%%%%%
\section{Conclusion}
We present a new QA model that operates over incomplete KB and text documents to answer open-domain questions, which yields consistent improvements over previous methods on the WebQSP benchmark with incomplete KBs.
The results show that (1) with the graph attention technique, we can efficiently and accurately accumulate question-related knowledge for each KB entity in one-pass of the KB sub-graph;
(2) our designed gating mechanisms could successfully incorporate the encoded entity knowledge while processing the text documents.
In future work, we will extend the proposed idea to other QA tasks with evidence of multimodality, e.g. combining with symbolic approaches for visual QA~\cite{gan2017vqs,mao2019neuro,hu2019language}.

\bibliography{acl2019}

\begin{thebibliography}{21}
\expandafter\ifx\csname natexlab\endcsname\relax\def\natexlab#1{#1}\fi

\bibitem[{Chen et~al.(2017)Chen, Fisch, Weston, and Bordes}]{chen2017reading}
Danqi Chen, Adam Fisch, Jason Weston, and Antoine Bordes. 2017.
\newblock \href {https://doi.org/10.18653/v1/P17-1171} {Reading wikipedia to
  answer open-domain questions}.
\newblock In \emph{Proceedings of the 55th Annual Meeting of the Association
  for Computational Linguistics, {ACL} 2017, Vancouver, Canada, July 30 -
  August 4, Volume 1: Long Papers}, pages 1870--1879.

\bibitem[{Gan et~al.(2017)Gan, Li, Li, Sun, and Gong}]{gan2017vqs}
Chuang Gan, Yandong Li, Haoxiang Li, Chen Sun, and Boqing Gong. 2017.
\newblock Vqs: Linking segmentations to questions and answers for supervised
  attention in vqa and question-focused semantic segmentation.
\newblock In \emph{ICCV}, pages 1811--1820.

\bibitem[{Haveliwala(2002)}]{haveliwala2002topic}
Taher~H Haveliwala. 2002.
\newblock Topic-sensitive pagerank.
\newblock In \emph{Proceedings of the 11th international conference on World
  Wide Web}, pages 517--526. ACM.

\bibitem[{Hu et~al.(2019)Hu, Rohrbach, Darrell, and Saenko}]{hu2019language}
Ronghang Hu, Anna Rohrbach, Trevor Darrell, and Kate Saenko. 2019.
\newblock Language-conditioned graph networks for relational reasoning.
\newblock \emph{arXiv preprint arXiv:1905.04405}.

\bibitem[{Kipf and Welling(2016)}]{kipf2016semi}
Thomas~N. Kipf and Max Welling. 2016.
\newblock \href {http://arxiv.org/abs/1609.02907} {Semi-supervised
  classification with graph convolutional networks}.
\newblock \emph{CoRR}, abs/1609.02907.

\bibitem[{Mao et~al.(2019)Mao, Gan, Kohli, Tenenbaum, and Wu}]{mao2019neuro}
Jiayuan Mao, Chuang Gan, Pushmeet Kohli, Joshua~B Tenenbaum, and Jiajun Wu.
  2019.
\newblock The neuro-symbolic concept learner: Interpreting scenes, words, and
  sentences from natural supervision.
\newblock \emph{arXiv preprint arXiv:1904.12584}.

\bibitem[{Mihaylov and Frank(2018)}]{mihaylov2018knowledgeable}
Todor Mihaylov and Anette Frank. 2018.
\newblock \href {https://aclanthology.info/papers/P18-1076/p18-1076}
  {Knowledgeable reader: Enhancing cloze-style reading comprehension with
  external commonsense knowledge}.
\newblock In \emph{ACL 2018}, pages 821--832.

\bibitem[{Miller et~al.(2016)Miller, Fisch, Dodge, Karimi, Bordes, and
  Weston}]{miller2016key}
Alexander~H. Miller, Adam Fisch, Jesse Dodge, Amir{-}Hossein Karimi, Antoine
  Bordes, and Jason Weston. 2016.
\newblock \href {http://aclweb.org/anthology/D/D16/D16-1147.pdf} {Key-value
  memory networks for directly reading documents}.
\newblock In \emph{Proceedings of the 2016 Conference on Empirical Methods in
  Natural Language Processing, {EMNLP} 2016, Austin, Texas, USA, November 1-4,
  2016}, pages 1400--1409.

\bibitem[{Min et~al.(2013)Min, Grishman, Wan, Wang, and
  Gondek}]{min2013distant}
Bonan Min, Ralph Grishman, Li~Wan, Chang Wang, and David Gondek. 2013.
\newblock Distant supervision for relation extraction with an incomplete
  knowledge base.
\newblock In \emph{Proceedings of the 2013 Conference of the North American
  Chapter of the Association for Computational Linguistics: Human Language
  Technologies}, pages 777--782.

\bibitem[{Seo et~al.(2016)Seo, Kembhavi, Farhadi, and
  Hajishirzi}]{seo2016bidirectional}
Minjoon Seo, Aniruddha Kembhavi, Ali Farhadi, and Hannaneh Hajishirzi. 2016.
\newblock Bidirectional attention flow for machine comprehension.
\newblock \emph{arXiv preprint arXiv:1611.01603}.

\bibitem[{Song et~al.(2018)Song, Zhang, Wang, and Gildea}]{song2018graph}
Linfeng Song, Yue Zhang, Zhiguo Wang, and Daniel Gildea. 2018.
\newblock A graph-to-sequence model for amr-to-text generation.
\newblock \emph{arXiv preprint arXiv:1805.02473}.

\bibitem[{Sun et~al.(2018)Sun, Dhingra, Zaheer, Mazaitis, Salakhutdinov, and
  Cohen}]{sun2018open}
Haitian Sun, Bhuwan Dhingra, Manzil Zaheer, Kathryn Mazaitis, Ruslan
  Salakhutdinov, and William~W. Cohen. 2018.
\newblock \href {https://aclanthology.info/papers/D18-1455/d18-1455} {Open
  domain question answering using early fusion of knowledge bases and text}.
\newblock In \emph{Proceedings of the 2018 Conference on Empirical Methods in
  Natural Language Processing, Brussels, Belgium, October 31 - November 4,
  2018}, pages 4231--4242. Association for Computational Linguistics.

\bibitem[{Veli{\v{c}}kovi{\'c} et~al.(2017)Veli{\v{c}}kovi{\'c}, Cucurull,
  Casanova, Romero, Lio, and Bengio}]{velivckovic2017graph}
Petar Veli{\v{c}}kovi{\'c}, Guillem Cucurull, Arantxa Casanova, Adriana Romero,
  Pietro Lio, and Yoshua Bengio. 2017.
\newblock Graph attention networks.
\newblock \emph{arXiv preprint arXiv:1710.10903}.

\bibitem[{Wang et~al.(2018)Wang, Yu, Guo, Wang, Klinger, Zhang, Chang, Tesauro,
  Zhou, and Jiang}]{wang2018r}
Shuohang Wang, Mo~Yu, Xiaoxiao Guo, Zhiguo Wang, Tim Klinger, Wei Zhang, Shiyu
  Chang, Gerry Tesauro, Bowen Zhou, and Jing Jiang. 2018.
\newblock R 3: Reinforced ranker-reader for open-domain question answering.
\newblock In \emph{Thirty-Second AAAI Conference on Artificial Intelligence}.

\bibitem[{Xiong et~al.(2017)Xiong, Zhong, and Socher}]{xiong2017dcn+}
Caiming Xiong, Victor Zhong, and Richard Socher. 2017.
\newblock \href {http://arxiv.org/abs/1711.00106} {{DCN+:} mixed objective and
  deep residual coattention for question answering}.
\newblock \emph{CoRR}, abs/1711.00106.

\bibitem[{Xu et~al.(2018)Xu, Wu, Wang, and Sheinin}]{xu2018graph2seq}
Kun Xu, Lingfei Wu, Zhiguo Wang, and Vadim Sheinin. 2018.
\newblock Graph2seq: Graph to sequence learning with attention-based neural
  networks.
\newblock \emph{arXiv preprint arXiv:1804.00823}.

\bibitem[{Yang and Mitchell(2017)}]{yang2017leveraging}
Bishan Yang and Tom Mitchell. 2017.
\newblock Leveraging knowledge bases in lstms for improving machine reading.
\newblock In \emph{Proceedings of the 55th Annual Meeting of the Association
  for Computational Linguistics (Volume 1: Long Papers)}, volume~1, pages
  1436--1446.

\bibitem[{Yih et~al.(2014)Yih, He, and Meek}]{yih2014semantic}
Wen-tau Yih, Xiaodong He, and Christopher Meek. 2014.
\newblock Semantic parsing for single-relation question answering.
\newblock In \emph{ACL 2014}, volume~2, pages 643--648.

\bibitem[{Yih et~al.(2016)Yih, Richardson, Meek, Chang, and Suh}]{yih2016value}
Wen-tau Yih, Matthew Richardson, Chris Meek, Ming-Wei Chang, and Jina Suh.
  2016.
\newblock The value of semantic parse labeling for knowledge base question
  answering.
\newblock In \emph{Proceedings of the 54th Annual Meeting of the Association
  for Computational Linguistics (Volume 2: Short Papers)}, volume~2, pages
  201--206.

\bibitem[{Yu et~al.(2018)Yu, Dohan, Luong, Zhao, Chen, Norouzi, and
  Le}]{yu2018qanet}
Adams~Wei Yu, David Dohan, Minh{-}Thang Luong, Rui Zhao, Kai Chen, Mohammad
  Norouzi, and Quoc~V. Le. 2018.
\newblock \href {http://arxiv.org/abs/1804.09541} {Qanet: Combining local
  convolution with global self-attention for reading comprehension}.
\newblock \emph{CoRR}, abs/1804.09541.

\bibitem[{Yu et~al.(2017)Yu, Yin, Hasan, dos Santos, Xiang, and
  Zhou}]{yu2017improved}
Mo~Yu, Wenpeng Yin, Kazi~Saidul Hasan, C{\'{\i}}cero~Nogueira dos Santos, Bing
  Xiang, and Bowen Zhou. 2017.
\newblock \href {https://doi.org/10.18653/v1/P17-1053} {Improved neural
  relation detection for knowledge base question answering}.
\newblock In \emph{Proceedings of the 55th Annual Meeting of the Association
  for Computational Linguistics, {ACL} 2017, Vancouver, Canada, July 30 -
  August 4, Volume 1: Long Papers}, pages 571--581.

\end{thebibliography}
\bibliographystyle{acl_natbib}

\end{document}